# Uncertainty-Aware Tightly-Coupled GPS Fused LIO-SLAM

Sabir Hossain
Faculty of Engineering and Applied Science,
Ontario Tech University, Canada
sabir.hossain@ontariotechu.net

Xianke Lin*
Faculty of Engineering and Applied Science,
Ontario Tech University, Canada
xiankelin@ieee.org

***Abstract*—Delivery robots aim to achieve high precision to facilitate complete autonomy. A precise three-dimensional point cloud map of sidewalk surroundings is required to estimate self-location. With or without the loop closing method, the cumulative error increases gradually after mapping for larger urban or city maps due to sensor drift. Therefore, there is a high risk of using the drifted or misaligned map. This article presented a technique for fusing GPS to update the 3D point cloud and eliminate cumulative error. The proposed method shows outstanding results in quantitative comparison and qualitative evaluation with other existing methods.**

## I. INTRODUCTION

Autonomous delivery vehicles (ADVs) provide a promising low-cost delivery solution. A growing number of ADV firms in the USA, such as AutoX, NURO and uDelv, has brought an influential impact on the delivery services market [1]. Most of these ADV technologies help boost the adoption of self-driving vehicle technology [2]. Moreover, ADVs are a technology that can be utilized for emergency shipment and supplementary during the period of world pandemics. A fundamental requirement for an autonomous delivery vehicle is accurate and robust mapping information for precise localization. The system requires a point-cloud-based map of urban and semi-indoor environments to achieve autonomy. Simultaneous localization and mapping (SLAM) is a crucial component in an autonomous delivery vehicle (ADV). Using ADV LiADR data and perception output, a SLAM module can estimate the location of an ADV preciously and build and update a 3D world map [3]. However, many of the existing mapping algorithms do not perform well in the case of long-distance mapping. In the case of mapping for ADVs, there is a need to use multiple routes that usually intersect with each other or reuse a route to obtain a complete map. In some situations, maps do not merge properly, generating a faulty map like Fig. 1. This translational drift confirms that only mapping based on LiDAR-inertial state estimation will not work properly in a large urban or city map. The translational drift happens in all three directions which make the map problematic. More significantly, not only the routes don't intersect, but also there will be a drift from original position in long range mapping. ADVs have to encounters with pedestrian, cyclist, intersections. Unreliable map can hardly be used for this purpose since there are many risks involved. Since LiDAR odometry has disadvantages such as the existence of accumulative errors and the dependence on shapes of obstacles in the scene. Final odometry of fusing LiDAR data and IMU data will be relatively error-free than the odometry from only LiDAR, but the accumulative error still exists for large-scale SLAM.

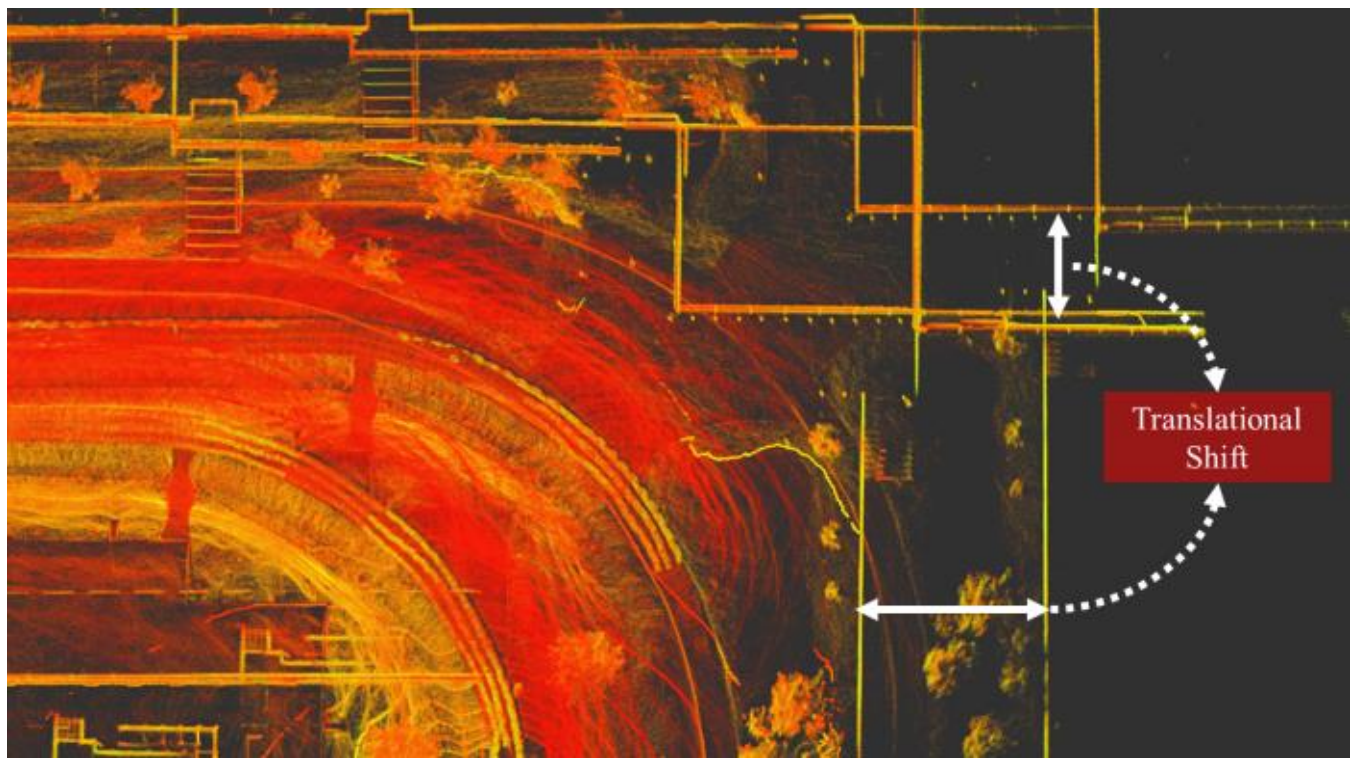

Fig. 1: Translational shift after a long range of mapping (Top View)

Fig. 1 shows translational drift after a long mapping period using FAST-LIO [4] algorithm, producing a faulty map. To overcome this limitation, including in z-directional shift, in this paper, we make the following contributions:

1. We propose a system that fuses GPS odometry with the LiDAR inertial odometry in a tightly-coupled manner. The proposed method utilizes the global positioning system to overcome the long-period drift issue in all three directions. Therefore, the method can generate a map with global positioning accuracy with multiple records/runs of similar places (return to the similar road, cross junction area)

2. The proposed method presents an uncertainty-aware logic-based system to fall back to the LIO system in a GPS-denied environment or unreliable GPS moments. During the time of GPS outage or unreliable GPS data, the method compensates for the state estimation from LIO. Therefore, it is a reliable mapping and state estimation tool for a large map that contains both GPS-accessible and GPS-denied environments.

3. The proposed method ensures the proper alignment of the LIO and GPS odometry frames using translational

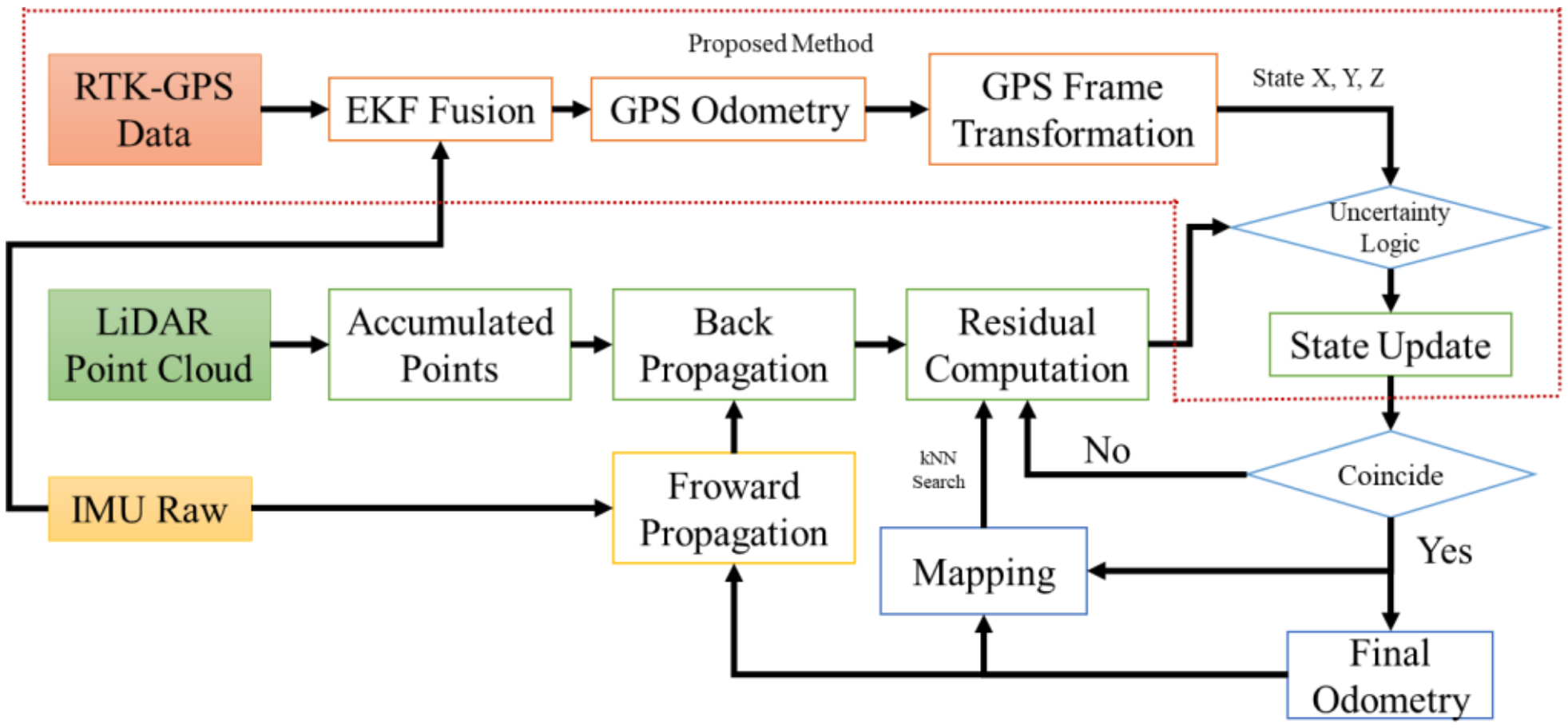

Fig. 2: Proposed System

and yaw extrinsic. Then, it utilizes the translational data from GPS odometry to generate a point cloud map without positional drift. The new data can be added later with GPS accuracy. Thus, our solution can produce accurate maps in both outdoor and semi-indoor/indoor environments.

The remainder of this paper is organized as follows. The “Related Studies” section describes a review of related studies. In the “Method” section, the proposed method is formulated. The “Result and Evaluation” section presents the experiments conducted to evaluate the proposed mapping method. Finally, conclusions are presented in the “Conclusion” section.

## II. Related Studies

A graph-based SLAM method aims to tag each point cloud with absolute GPS position information [5]. This method provides an accurate map for localization, and the points clouds are apprised with absolute GPS coordinate information. However, GPS instability is not taken into consideration while stamping absolute coordinates for each point cloud. SLAM based on GPS and LIDAR is introduced for outdoor firefight robots [6] (specific mapping for petrochemical buildings). This technique did not take inertial propagation into account while mapping. LOAM is introduced by revolving 2D LiDAR for odometry and mapping [7]. However, the optimization process of the algorithm does not require IMUs.

GLO-SLAM [8] relies on a lightweight backend algorithm to overcome the accumulative error in real-time. The proposed VGL algorithm confirm the reliability of GPS message. Also, their proposed action against GPS instability covers the drift in one single plane (XY). Similarly, LiDAR-inertial localization and GPS optimization are used for mapping and state estimation using the GNSS module [9]. Moreover, GPS-optimized odometry for the online state estimator was running at around 1Hz, which is very low. Our proposed method evaluates the reliability of RTK-GPS from the uncertainty covariance in all 3D planes. Moreover, the LiDAR odometry update is from Fast-LIO [10], a computationally efficient LIO module that uses iterated KF for optimization and parallel KD-tree to improve computational loss [4]. Therefore, the positional update is very accurate when the module is running indoors.

LeGO-LOAM [11] is a lightweight and optimized mapping tool that fuses LOAM and IMU. Similarly, LiDAR, IMU factor and GPS factor(optional) is fused in a loosely-coupled manner using EKF in LIO-SAM [12] mapping. The GPS factor in LIO-SAM optimizes lidar odometry in the factor graph [12]. However, this mapping approach relies heavily on IMU calibration and does not provide results for the z-direction drift. On the other hand, GPS fusion is used to reduce the accumulative error for visual-SLAM modules [13]. LiDAR-based real-time state estimation is proposed [14] for a drone in GPS-denied environments. Since our approach aims to solve the mapping and localization problem of autonomous delivery, the solution mainly needs to solve outdoor situations. Inertial propagation and GPS component are used in the proposed method. Most algorithms did not cover the limitations of GPS instability and altitude consideration. Our approaches proposed the method considering these factors to overcome mismatching in the intersection, end-to-end translation error and multi-run drift.

## III. Method

The proposed pipeline consists of state estimation and logic-based state update that is finally provided to the mapping module. First, GPS and IMU data are fused using EKF, shown in Fig. 2. Then, the GPS Odom data is transformed into the robot sensor frame in order to match the heading. Finally, the translational state (x, y, z) goes for the state update module through the uncertainty logic module.

### *A. GPS Fusion*

Centimeter-level localization accuracy can be achieved with a very accurate global navigation satellite system (GNSS), such as a real-time kinematic (RTK) module. Carrier-phase location information from base stations is wirelessly transmitted to the RTK device to rectify the estimated error in real-time. The proposed method tightly depends on GPS position. RTK and IMU sensors are transformed and fused using EKF that produces the GPS odometry. The odometry is then transformed into the LiDAR frame. Only states X, Y, and Z are used for state update (Shown in the figure)

### B. *EKF Fusion and Frame Transformation*

Since the EKF robot localization is primarily employed as an odometry filter, it includes differential position estimations that include some absolute position estimation. This module works by initializing the pose node that integrates the filtered GPS for absolute position and IMU measurement [15]. Because of tall buildings, trees, or the unavailability of GPS-fix, only GPS-based navigation will be a complete disaster. The frequent lack of GPS-fix will cause the GPS to provide wrong values. Surprisingly resilient to such failures, at least for a short period of time, EKF-based estimations fused with other sensors are the solution. Without any sensor directly reporting the desired state variable, EKF can also estimate location. The magnetic heading tends to produce irregular data in the presence of motors and batteries in the delivery vehicle. Improperly isolated USB-3 cables, which can be used for other sensors, can cause GPS-fix disruption. As a result, the robot effectively lacks proper state estimation in the world frame. EKF can still assist in such cases in obtaining a reliable estimate of non-measurable state variables. The delivery vehicle is equipped with rotating LiDAR, IMU and GPS devices that must maintain the requirements:

1. Sensors in delivery vehicles typically use the ENU frame. The same frame should be used for both the GPS and IMU measurements.
2. The RTK GPS data should include the GPS measurement, which will appear as an odometry message. This also gives a covariance estimate for the measurement. The covariance on the orientation and twist message should be absurdly large to ensure that the filter ignores this data because GPS is a poor source for orientation information.
3. Since KF offers smooth filtered odometry from the GPS data, the GPS must have a reasonably strong signal in order for the EKF output odometry to be accurate.

The module was designed [15] to fuse as many as sensors and track 15 state values for the robot, which are – $X, Y, Z, roll, pitch, yaw, \dot{X}, \dot{Y}, \dot{Z}, \dot{roll}, \dot{pitch}, \dot{yaw}, \ddot{X}, \ddot{Y}, \ddot{Z}$. The EKF estimates the 6-DOF pose and velocity. The prediction step of the algorithm consists of equation (1) for the output of the current state and equation (2) for error covariance in time:

$$\hat{x}_k = f(\boldsymbol{x}_{k-1}) \tag{1}$$

$$\widehat{\boldsymbol{P}}_{\boldsymbol{k}} = \boldsymbol{Q} + \boldsymbol{F}^T \boldsymbol{P}_{k-1} \boldsymbol{F} \tag{2}$$

Here, The estimate error covariance, $\boldsymbol{P}$, is projected via $\boldsymbol{F}$, the Jacobian of $f$, and then perturbed by the process noise covariance $\boldsymbol{Q}$. These state estimation equations are adopted from [15]. The correction step measures the Kalman gain from the observation matrix, $\boldsymbol{H}$ and measurement covariance $P_k$ that updates the state vector and covariance matrix. Joseph form covariance is used to update the equation to promote filter stability by ensuring that $\boldsymbol{P_k}$ remains positive semi-definite. The correction step is as follows from equation (3)-(5):

$$K = H^T \hat{P}_k \left(R + H^T \hat{P}_k H\right)^{-1} \tag{3}$$

$$x_k = K(z - \hat{x}_k H) + \hat{x}_k \tag{4}$$

$$P_k = K^T R K + (I - KH)^T \hat{P}_k (I - KH) \tag{5}$$

In order to use GPS with the EKF, the GPS message needs to be transformed to the robot frame, in simplest terms, in GPS odometry. The measurement equation (6) is given below [15]:

$$\begin{bmatrix} x_{\mathrm{odm}} \\ y_{\mathrm{odm}} \\ z_{\mathrm{odm}} \\ 1 \end{bmatrix} = \boldsymbol{T}^{-1} \begin{bmatrix} x_{utm_t} \\ y_{utm_t} \\ z_{utm_t} \\ 1 \end{bmatrix} \tag{6}$$

where $x_{\mathrm{utm_t}}, y_{\mathrm{utm_t}}$, and $z_{\mathrm{utm_t}}$ are the UTM coordinates at subsequent $t$ time and $x_{\mathrm{odm}}$, $y_{\mathrm{odm}}$ and $z_{\mathrm{odm}}$ are the output from the GPS Odom measurement. $\boldsymbol{T}^{-1}$ is the inverse of the transformation matrix, which transforms the robot frame to UTM coordinates [15] shown in the following equation (7) :

$$\boldsymbol{T}^{-1} = \begin{bmatrix} \boldsymbol{c}\theta\boldsymbol{c}\psi & \boldsymbol{c}\psi\boldsymbol{s}\varphi\boldsymbol{s}\theta - \boldsymbol{c}\varphi\boldsymbol{s}\psi & \boldsymbol{c}\varphi\boldsymbol{c}\psi\boldsymbol{s}\theta + \boldsymbol{s}\theta\boldsymbol{s}\psi & x_{\mathrm{utm_0}} \\ \boldsymbol{c}\theta\boldsymbol{s}\psi & \boldsymbol{c}\varphi\boldsymbol{c}\psi + \boldsymbol{s}\varphi\boldsymbol{s}\theta\boldsymbol{s}\psi & -\boldsymbol{c}\psi\boldsymbol{s}\varphi + \boldsymbol{c}\varphi\boldsymbol{s}\theta\boldsymbol{s}\psi & y_{\mathrm{utm_0}} \\ -sin\theta & cos\theta sin\varphi & cos\varphi sin\theta & z_{\mathrm{utm_0}} \\ 0 & 0 & 0 & 1 \end{bmatrix}^{-1} \tag{7}$$

where $\varphi, \theta$, and $\psi$ are the vehicle's initial UTM-frame roll, pitch, and yaw, respectively. $\boldsymbol{c}$ and $\boldsymbol{s}$ designate the cosine and sine functions, respectively, and $x_{\mathrm{UTM_0}}, y_{\mathrm{UTM}}$, and $z_{\mathrm{UTM_0}}$ are the UTM coordinates of the first reported GPS position.

### C. *LIO-SLAM*

The proposed pipeline is developed over the FAST-LIO2 algorithm [4], which is open source under the name of FAST-LIO. FAST-LIO2 uses the iterative kd-tree search for parallel computation and fast iterated KF for odometry optimization. In Fast-LIO [10], edge features are produced from the feature extraction module, which only takes LiDAR data as input. The state estimation system utilizes the extracted features (planer and edge feature) and IMU to estimate the position. The feature points are registered to the global frame and then combined with the feature map continuously from the updated pose estimation. However, instead of feature extraction, Fast-LIO2 accumulated the LiDAR points at high frequency. The iterated KF framework is passed through this large local map in a tightly coupled manner to calculate the optimized odometry. The incremental k-d tree structure assembles the global map points parallelly in the local map. The unnecessary points from the ikd-tree get removed based on the FoV threshold provided. New points are resisted with respect to the optimized pose from iterative KF and produce a unified map from the iKD-tree. For clarification, in this paper, Fast-LIO is predominantly referring Fast-LIO2. The mapping portion in the Fig. 2 is actually real-time mapping with ikd-support. On-tree downscaling is used for point-wise interstation outside of the data structure. Box-wise deletion is performed on unnecessary points during the tree rebuild process.

### *D. Uncertainty Logic Calculation*

Logic-based decision approach is adopted, which depends on uncertainty calculation. The covariance matrix is diagonal. The variance of each state in the state vector can be obtained from the diagonal elements of the covariance matrix. Due to defective sensors, the sensor message uses a position of available data. Similarly, the device will fail to measure covariance properly if GPS drops out. The filter's covariance matrix maintains its stability and the variance values despite the significant discrepancy between the state estimate and measurement. Therefore, the covariance matrix is a notifier for the instability of each value in the state vector. Noise covariance is obtained from the GPS odometry output from the localization module. The algorithm I show the decision function of calculating state update based on uncertainty -

Algorithm I

**Input**: fusion.state update, LIO.state.update, noise.uncertainty
**Output**: state.update
Threshold: Set a constant

**Function Start**
If noise. uncertainty[x] > Threshold[x]:
State.update.x ← LIO.state.update.x
else
State.update.x ← fusion.state.update.x

If noise. uncertainty[y] > Threshold[y]:
State.update.x ← LIO.state.update.y
else
State.update.x ← fusion.state.update.y

If noise.uncertainty[z] > Threshold[z]:
State.update.z ← LIO.state.update.z
else
State.update.z ← fusion.state.update.z
**Function End**

The covariance matrix values for the pose state increase when the uncertainty increases (shown in Fig. 3). A threshold level is set for each of them to obtain a more robust state update result for mapping.

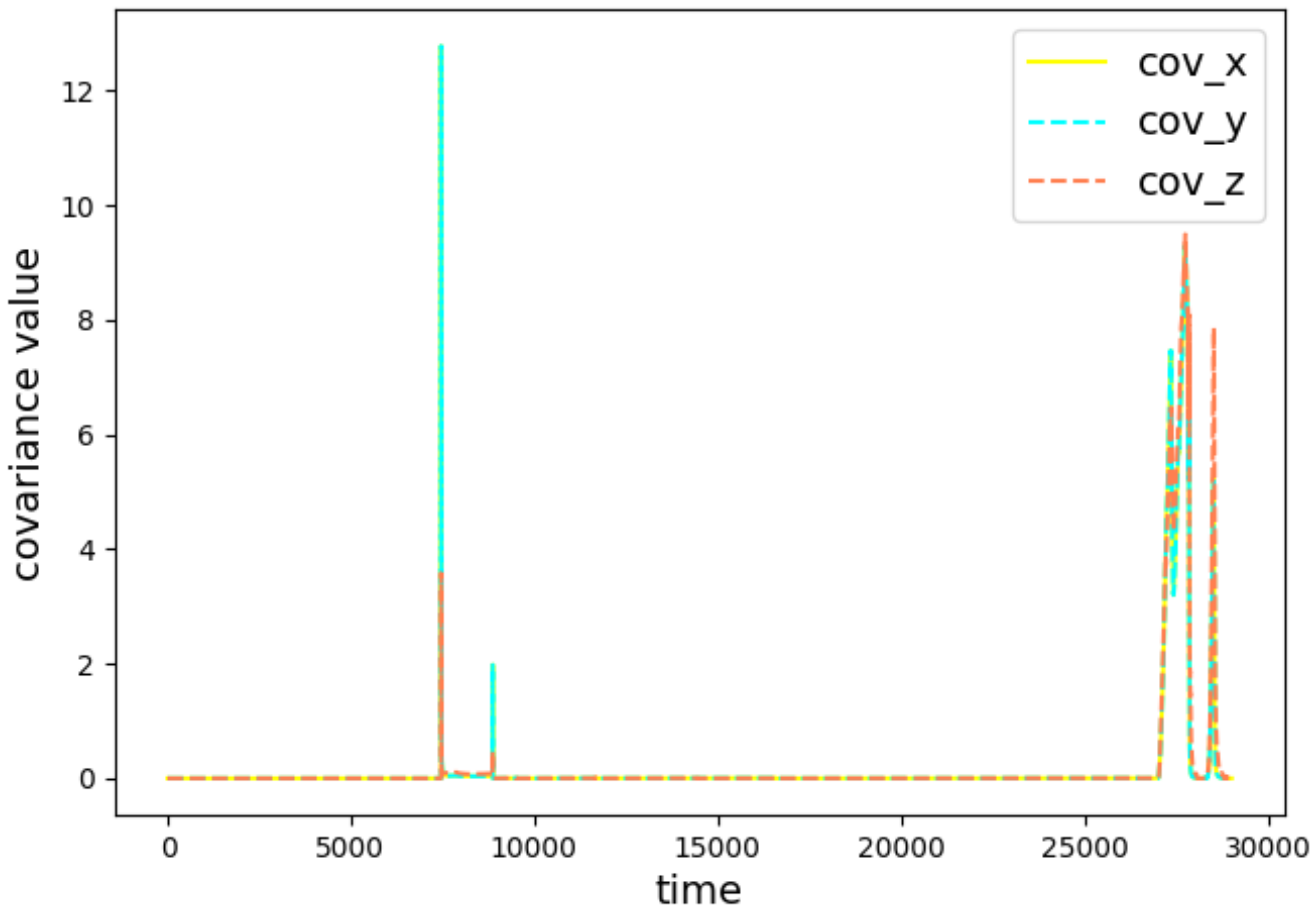

Fig. 3: Increase in uncertainty during an experiment

## IV. RESULTS & EVALUATION

To verify the proposed algorithm, the framework is processed in two rosbag recorded on different days (named experiments 1 and 2) at Ontario Tech University campus (partially neighbourhood). The produced point map using our approach is aligned with the google earth map over the campus (shown in Fig 4 and Fig.5). Fig.6 (a), (b), and (c) show the qualitative visual result of translational drift in a long-range dataset. Our approach produces a robust and accurate map overcoming those drifts (shown in Fig.6). Complete and enlarged trajectory results for both of our recorded datasets (experiment 1 and 2) are shown in Fig. 7 and Fig. 8, respectively. Fig. 7(b) and Fig.8(b) marked the RTK dropout location that is inside the building, producing a very unreliable state estimation. This limitation is resolved by using an uncertainty measurement taken into account, shown in Fig. 7(b). It increases uncertainty at the end of the dataset, like Fig. 3. Similarly, Fig.8 (b) shows a large end-to-end error and Fig.8 (c) shows a multi-run translational error. Fig. 11 presents a 3D plot comparison of GPS, Fast-LIO, and GPS-fused LIO with uncertainty (our approach). It is the 3d version of Fig. 8(a). It contains an uncertainty graph that indicates GPS dropout three times. It shows the full trajectory path in 3-dimensional space with z-directional drift for Fast-LIO in a 3D plot and three spikes in uncertainty three times in the whole experiment. Our approach overcomes this multi-directional drift. For clarification, LIO-FUSION refers to our approach without uncertainty logic, LIO-FUSION-UNCERTAINTY refers to the complete proposed framework with uncertainty, the yellow highlighted mark refers to GPS-dropout/denied situation, and Fast-LIO predominantly refers to Fast-LIO2 [4] in the result section.

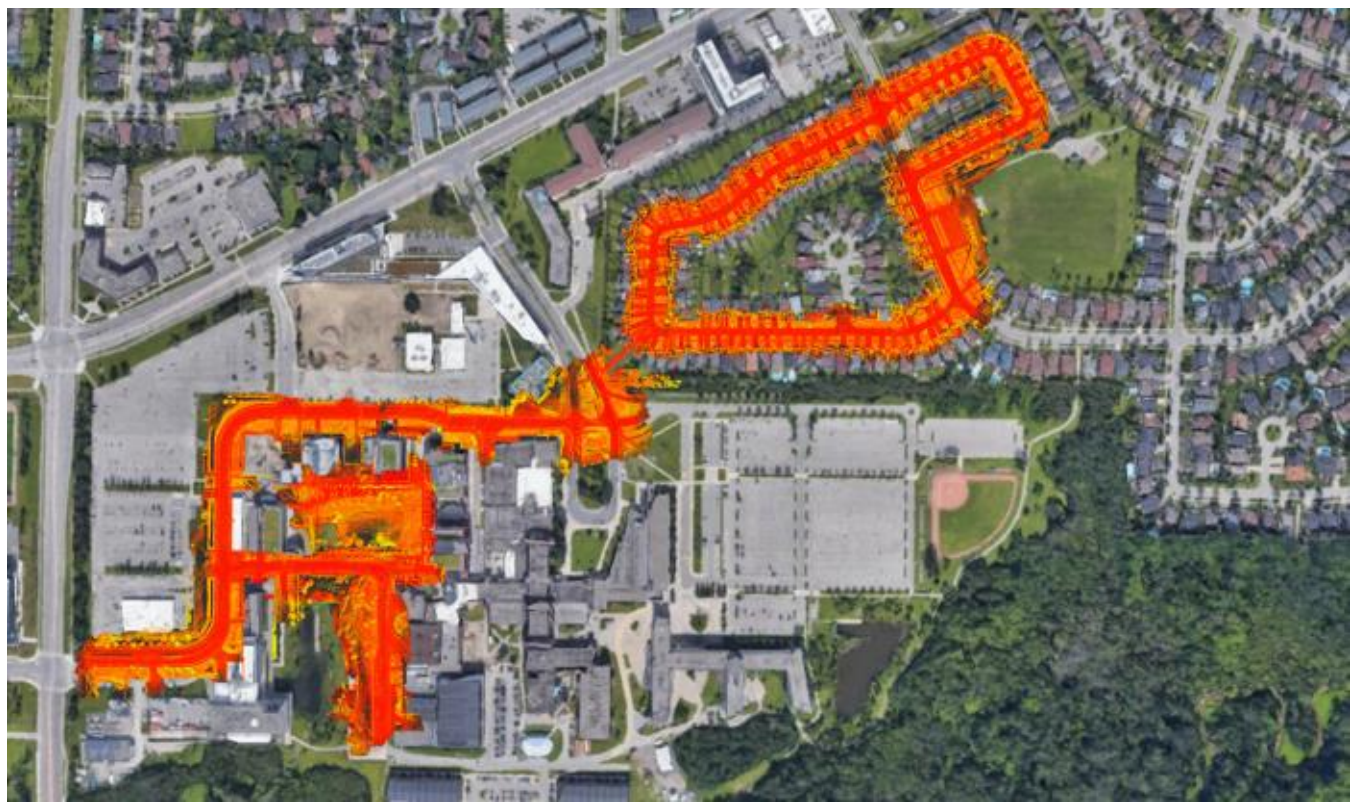

Fig. 4: PCL-map from 1st experiment aligned to Google Earth Map (Bird’s Eye View)

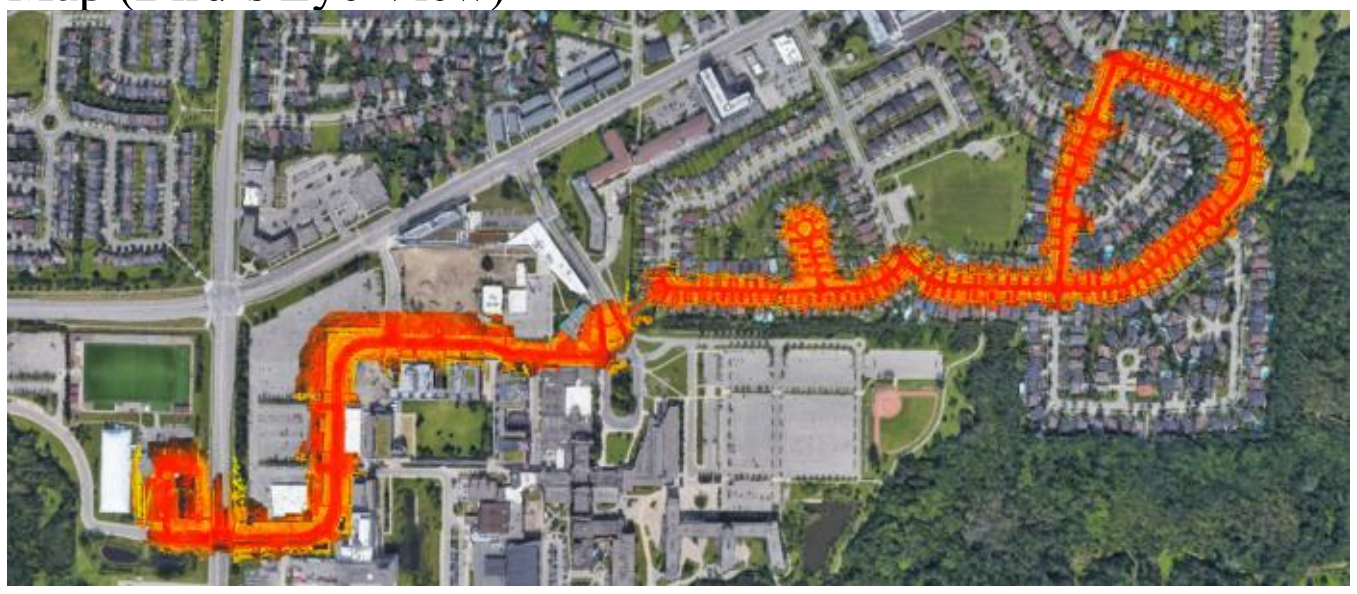

Fig. 5: PCL-map from 2nd experiment aligned to Google Earth Map (Bird’s Eye View)

Furthermore, quantitative evaluation is performed to compare our approach with the existing method LIO-SAM [12], which uses GPS factor as an input. Fig. 9 and Fig. 10 present the qualitative trajectory evaluation result for the park and small campus datasets, respectively. Table I shows a quantitative comparison of mapping approaches. Our approach (LIO-FUSION with uncertainty) achieved less RMSE value with respect to GPS for the small campus and park dataset that LIO-SAM provided.

Table I: Translational RMS error w.r.t GPS Odom

| Dataset | LIO-SAM [12] | FAST-LIO [4] | Our Approach |
|---|---|---|---|
| Small Campus | 0.59 | 0.62 | 0.18 |
| Park | 0.54 | 0.94 | 0.12 |

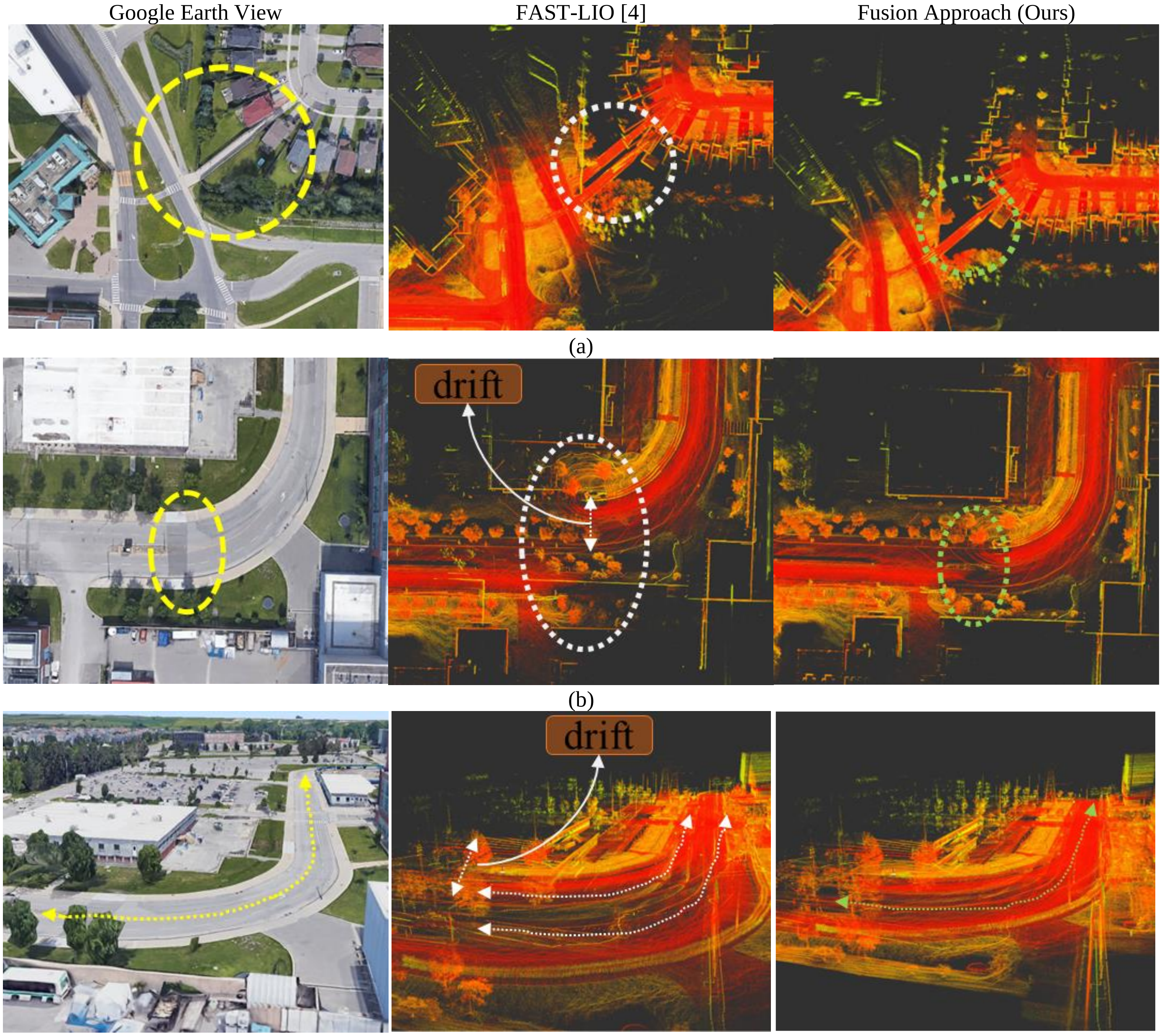

(c)

Fig. 6: GPS fusion capable of overcoming long-range translational drift (a) multi-run drift error in Fast-LIO approach; GPS fused approach with no drift on the same location shown in google earth image (Bird's Eye View) (b) major end-to-end translation error in Fast-LIO approach after long range mapping and aligned map overcoming large end-to-end error at the same location shown in google map (Bird's Eye View) (c) major end-to-end translation error including z-direction using Fast-LIO and robust map without end-to-end error at the same location shown in the google earth image (3D View)

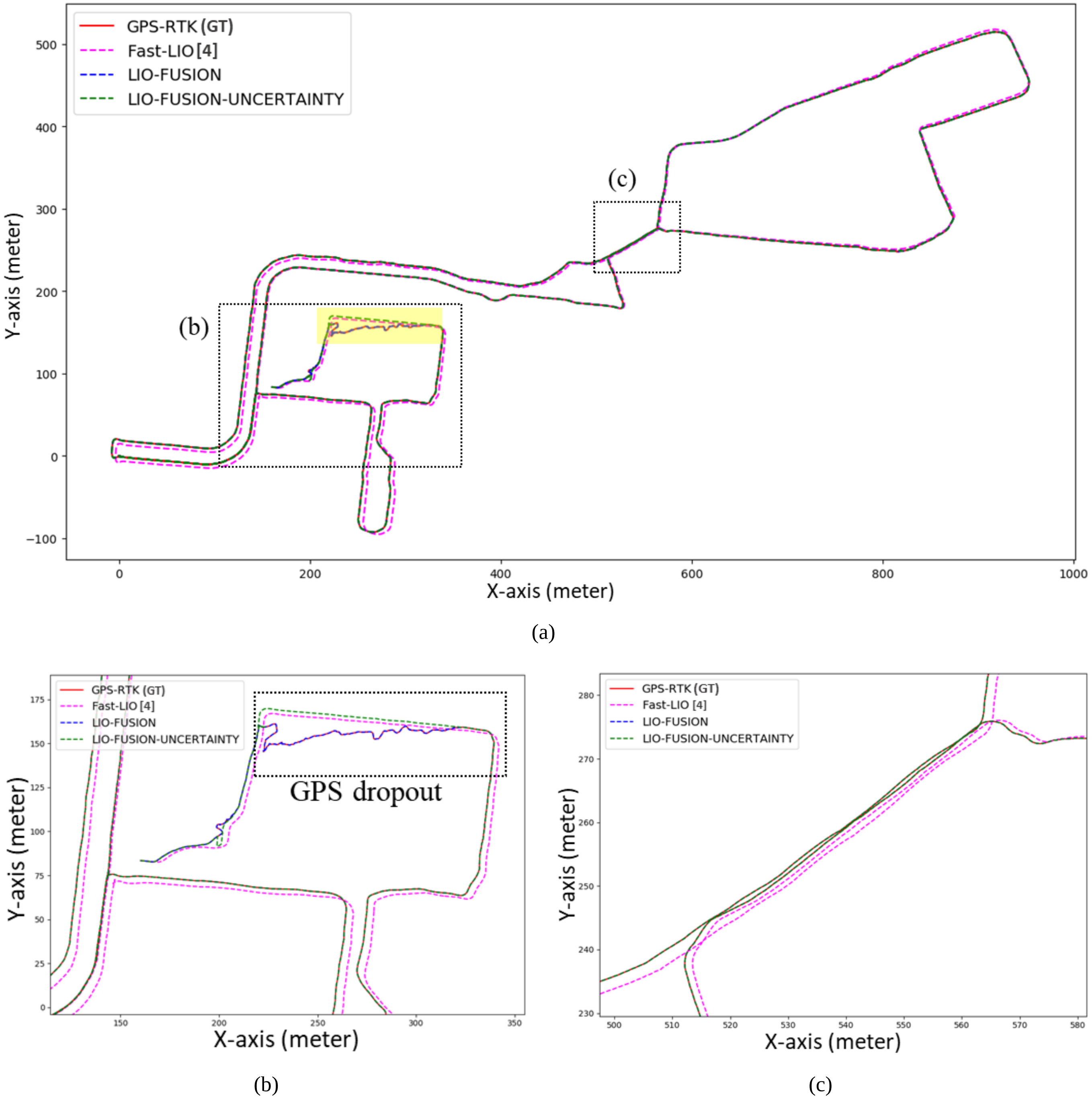

Fig. 7: GPS fusion result on our dataset (experiment 1); Comparison between GPS-RTK, Fast-LIO, GPS-fused LIO and GPS-fused LIO with uncertainty (a) fully mapping trajectory, (a) fully mapping trajectory, (b) Capable of working both in indoor and outdoor during the time of dropout, (c) translational drift on multi-run path showing long-range position drift is overcome by GPS-fusion

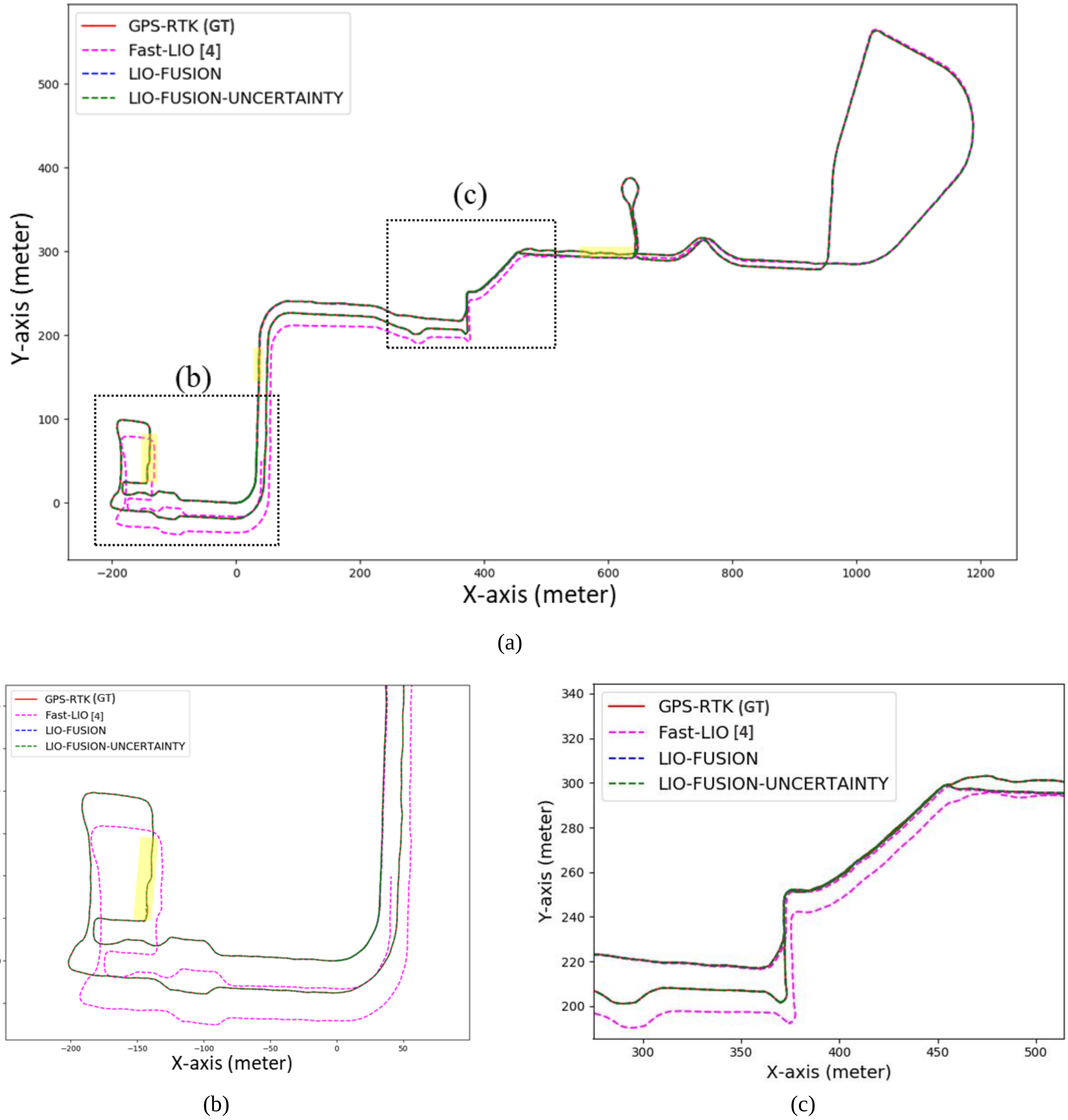

Fig. 8: GPS fusion result on our dataset (experiment 2); Comparison between GPS-RTK, Fast-LIO [4], GPS-fused LIO and GPS-fused LIO with uncertainty (a) fully mapping trajectory, (b) Capable of working both in indoor and outdoor during the time of dropout, (c) translational drift on multi-run path showing long-range position drift is overcome by GPS-fusion

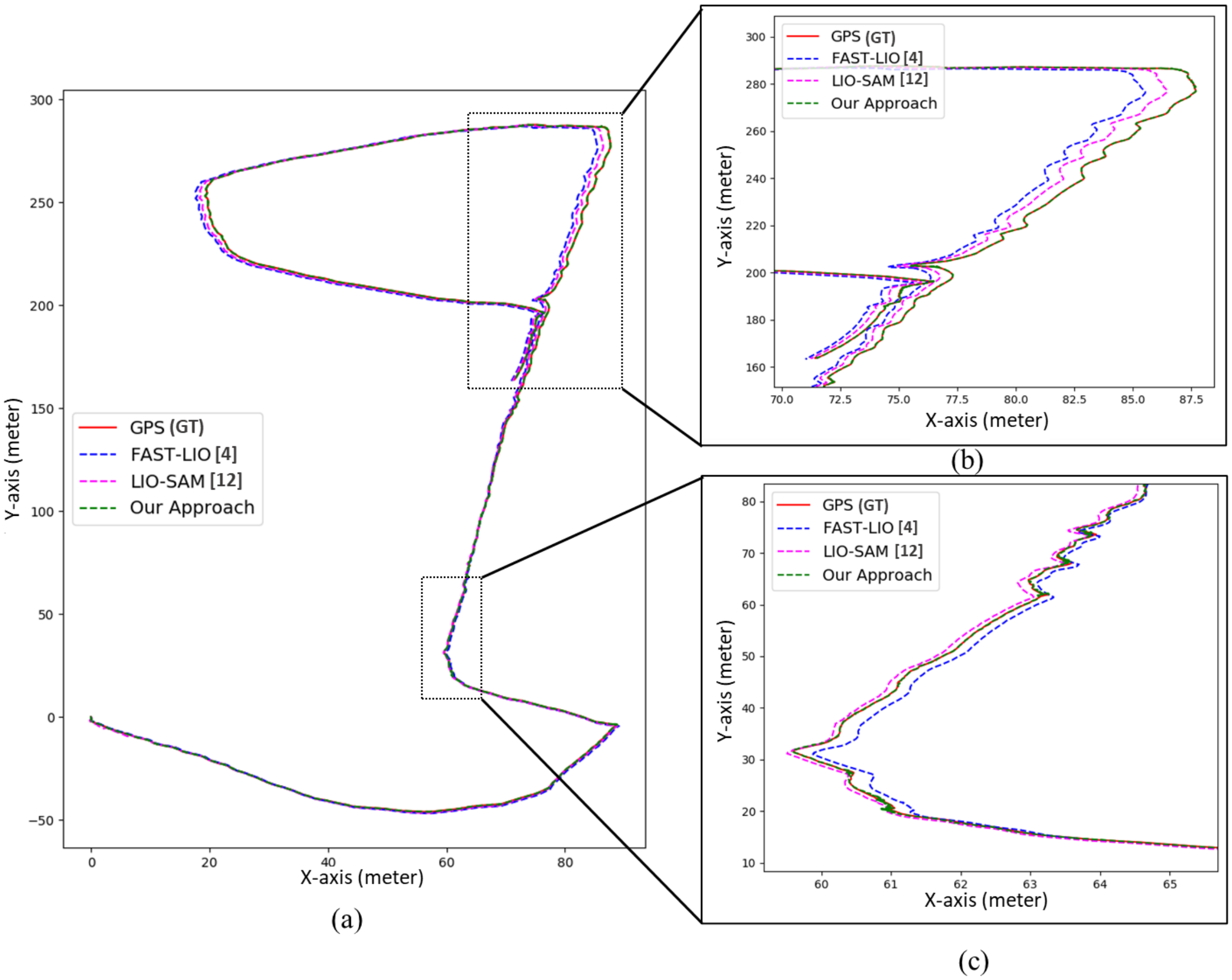

Fig. 9: Trajectory comparison of GPS, Fast-LIO [4], LIO-SAM (includes all factors) [12] and our approach; (a) Full path of Park Dataset provided by LIO-SAM authors; (b), (c), (d) shows an enlarged portion of the full map

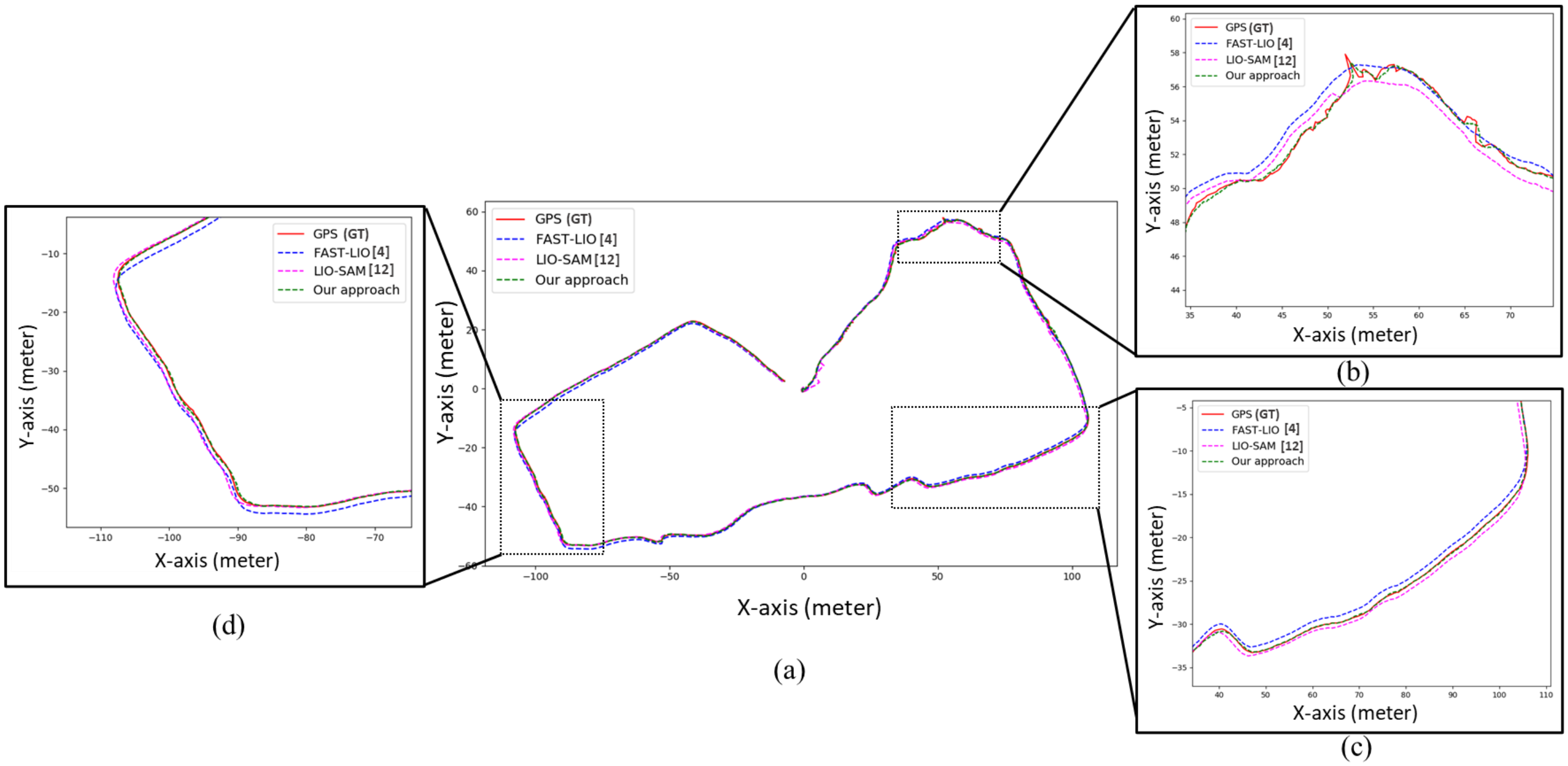

Fig. 10: Trajectory comparison of GPS, Fast-LIO [4], LIO-SAM (includes all factors) [12] and our approach; (a) Full path of Small Campus Dataset provided by LIO-SAM authors; (b), (c), (d) shows an enlarged portion of the full map

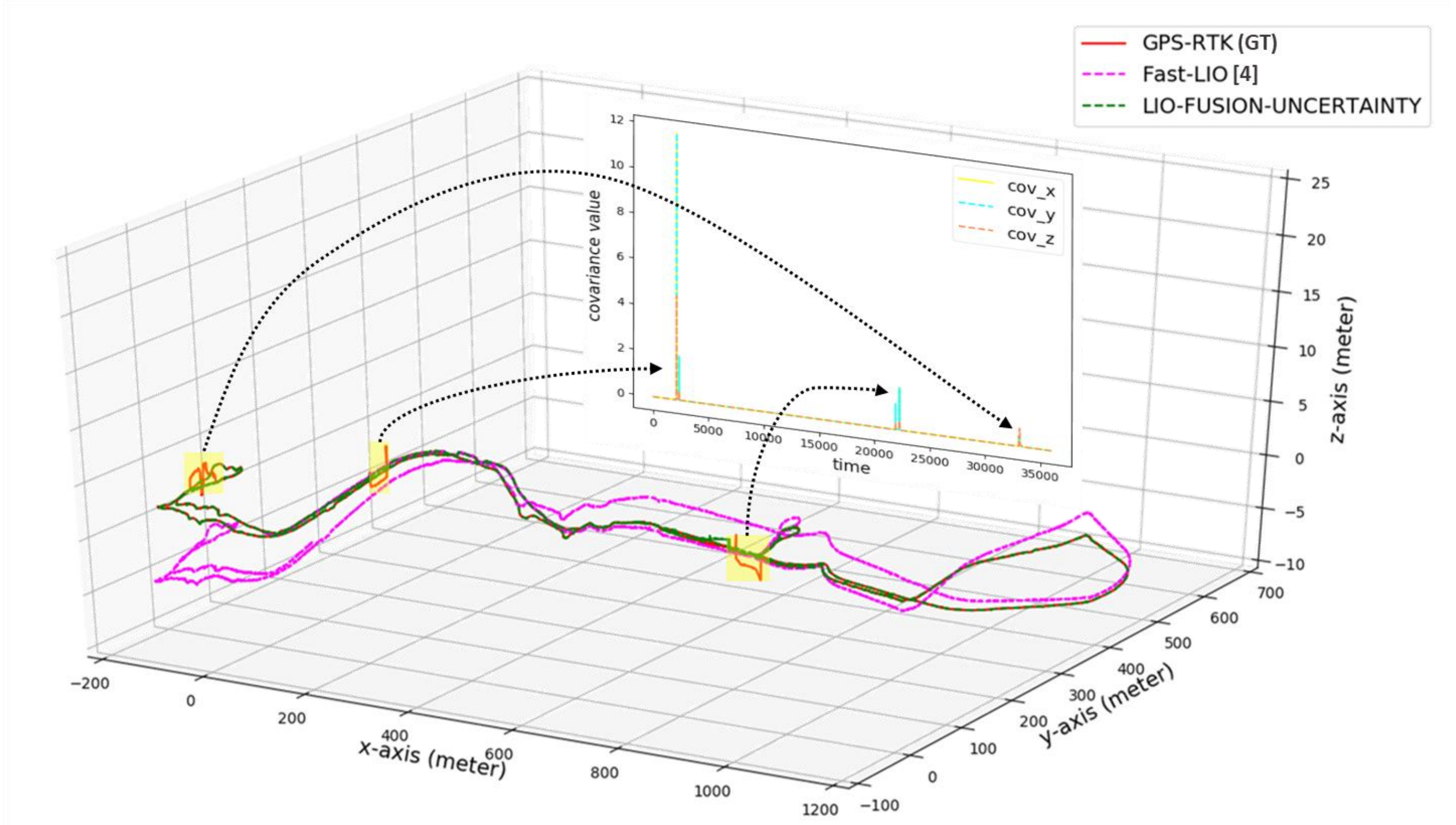

Fig. 11: Trajectory comparison of GPS, Fast-LIO, and GPS-fused LIO with uncertainty (our approach); three spikes (highlighted yellow) of unreliable GPS data in the whole experiment; Full trajectory showing z-directional drift for Fast-LIO in 3D plot; Our approach without multi-directional drift (including z)

## V. Conclusion

The proposed method is a feasible solution for mapping an autonomous delivery vehicle that will operate in areas consisting of GPS-accessed(outdoor) and GPS-denied (partially indoor). In this paper, we fuse GPS odometry with the LiDAR inertial odometry in a tightly-coupled manner. The final state estimation is the fusion of GPS and IMU using EKF. Since the proposed method utilizes the global positioning system, it can overcome the long-range drift issue in all three directions, including z-direction. Additionally, the proposed method presents an uncertainty-aware logic-based system that switches the positional update back to LIO-system in case of a GPS-denied situation. Then, it utilizes the final translational data from the fusion to generate a point cloud map without positional drift. Thus, our solution can produce accurate maps in outdoor and semi-indoor/indoor environments. Results on our dataset verify the mapping approach and results on other datasets (park and small campus) display accuracy for mapping. The quantitative comparison for LIO-SAM is performed since this existing mapping approach also uses the GPS factor in their mapping approach. Our approach (LIO-FUSION with uncertainty) achieved less RMSE value with respect to GPS for the small campus and park dataset that LIO-SAM provided.

The GPS localization can be improved in future work by integrating it with visual-SLAM modules [16] [17]. Furthermore, a filter can be used during the time of logic change in order to achieve a smooth transition in position. Therefore, it will produce more accurate positioning information during the time of dropout.

## Conflict of Interest

The authors declare no conflict of interest.

## Acknowledgment

The work is funded by LinLAB's start-up fund.

## Declarations

No humans and/or animals were involved in performing this research.

## References

[1] M. Figliozzi and D. Jennings, “Autonomous delivery robots and their potential impacts on urban freight energy consumption and emissions,” *Transp. Res. procedia*, vol. 46, pp. 21–28, 2020.

[2] D. Jennings and M. Figliozzi, “Study of road autonomous delivery robots and their potential effects on freight efficiency and travel,” *Transp. Res. Rec.*, vol. 2674, no. 9, pp. 1019–1029, 2020, doi: 10.1177/0361198120933633.

[3] W. Wang, W. Zhao, X. Wang, Z. Jin, Y. Li, and T. Runge, “A low-cost simultaneous localization and mapping algorithm for last-mile indoor delivery,” in *2019 5th International Conference on Transportation Information and Safety (ICTIS)*, 2019, pp. 329–336.

[4] W. Xu, Y. Cai, D. He, J. Lin, and F. Zhang, "Fast-lio2: Fast direct lidar-inertial odometry," *IEEE Trans. Robot.*, 2022.

[5] J. Matsuo, K. Kondo, T. Murakami, T. Sato, Y. Kitsukawa, and J. Meguro, "3D Point Cloud Construction with Absolute Positions Using SLAM based on RTK-GNSS," in *The Abstracts of the international conference on advanced mechatronics : toward evolutionary fusion of IT and mechatronics : ICAM*, 2021, vol. 2021.7, no. 0, pp. GS7-2, doi: 10.1299/jsmeicam.2021.7.gs7-2.

[6] A. U. Shamsudin *et al.*, "Correction to: Consistent map building in petrochemical complexes for firefighter robots using SLAM based on GPS and LIDAR (ROBOMECH Journal, (2018), 5, 1, (7), 10.1186/s40648-018-0104-z)," *ROBOMECH J.*, vol. 5, no. 1, pp. 1–13, 2018, doi: 10.1186/s40648-018-0106-x.

[7] J. Zhang and S. Singh, "LOAM: Lidar Odometry and Mapping in Real-time," in *Robotics: Science and Systems*, 2014, vol. 2, no. 9, pp. 1–9, doi: 10.15607/RSS.2014.X.007.

[8] R. Lin, J. Xu, and J. Zhang, "GLO-SLAM: a slam system optimally combining GPS and LiDAR odometry," *Ind. Rob.*, vol. 48, no. 5, pp. 726–736, 2021, doi: 10.1108/IR-12-2020-0272.

[9] L. Zheng, Y. Zhu, B. Xue, M. Liu, and R. Fan, "Low-Cost GPS-Aided LiDAR State Estimation and Map Building," in *IST 2019 - IEEE International Conference on Imaging Systems and Techniques, Proceedings*, 2019, pp. 1–6, doi: 10.1109/IST48021.2019.9010530.

[10] W. Xu and F. Zhang, "FAST-LIO: A Fast, Robust LiDAR-inertial odometry package by tightly-coupled iterated kalman filter," *IEEE Robot. Autom. Lett.*, vol. 6, no. 2, pp. 3317–3324, 2021, doi: 10.1109/LRA.2021.3064227.

[11] T. Shan and B. Englot, "LeGO-LOAM: Lightweight and Ground-Optimized Lidar Odometry and Mapping on Variable Terrain," in *IEEE International Conference on Intelligent Robots and Systems*, 2018, pp. 4758–4765, doi: 10.1109/IROS.2018.8594299.

[12] T. Shan, B. Englot, D. Meyers, W. Wang, C. Ratti, and D. Rus, "LIO-SAM: Tightly-coupled lidar inertial odometry via smoothing and mapping," in *IEEE International Conference on Intelligent Robots and Systems*, 2020, pp. 5135–5142, doi: 10.1109/IROS45743.2020.9341176.

[13] S. Boche, X. Zuo, S. Schaefer, and S. Leutenegger, "Visual-Inertial SLAM with Tightly-Coupled Dropout-Tolerant GPS Fusion," *arXiv Prepr. arXiv2208.00709*, 2022, [Online]. Available: http://arxiv.org/abs/2208.00709.

[14] H. SHEN, Q. ZONG, H. LU, X. ZHANG, B. TIAN, and L. HE, "A distributed approach for lidar-based relative state estimation of multi-UAV in GPS-denied environments," *Chinese J. Aeronaut.*, vol. 35, no. 1, pp. 59–69, 2022, doi: 10.1016/j.cja.2021.04.021.

[15] T. Moore and D. Stouch, "A generalized extended Kalman filter implementation for the robot operating system," in *Advances in Intelligent Systems and Computing*, vol. 302, Springer, 2016, pp. 335–348.

[16] N. H. K. Tran and V. H. Nguyen, "An EKF-Based Fusion of Visual-Inertial Odometry and GPS for Global Robot Pose Estimation," in *2021 6th IEEE International Conference on Recent Advances and Innovations in Engineering, ICRAIE 2021*, 2021, vol. 6, pp. 1–5, doi: 10.1109/ICRAIE52900.2021.9703965.

[17] S. A. Berrabah, Y. Baudoin, and H. Sahli, "Combined GPS/INS/WSS integration and visual SLAM for geo-localization of a mobile robot," in *2011 IEEE International Conference on Mechatronics, ICM 2011 - Proceedings*, 2011, pp. 499–503, doi: 10.1109/ICMECH.2011.5971337.